\documentclass{article}

\usepackage{microtype}
\usepackage{graphicx}
\usepackage{subfigure}
\usepackage{booktabs} 
\usepackage{enumitem}

\usepackage{hyperref}
\usepackage{multirow}
\usepackage{amsmath}
\usepackage{amsfonts}
\usepackage{booktabs}
\usepackage{bm}
\usepackage{widetext}
\usepackage[ruled,vlined]{algorithm2e}

\DeclareMathOperator{\tr}{tr}

\usepackage[accepted]{icml2018}

\icmltitlerunning{Mean Field Theory of Recurrent Neural Networks}

\begin{document}

\twocolumn[
\icmltitle{Dynamical Isometry and a Mean Field Theory of RNNs:\\ Gating Enables Signal Propagation in Recurrent Neural Networks}


\icmlsetsymbol{equal}{*}

\begin{icmlauthorlist}
\icmlauthor{Minmin Chen}{equal,goog}
\icmlauthor{Jeffrey Pennington}{equal,brain}
\icmlauthor{Samuel S. Schoenholz}{equal,brain}
\end{icmlauthorlist}

\icmlaffiliation{goog}{Google}
\icmlaffiliation{brain}{Google Brain}

\icmlcorrespondingauthor{Minmin Chen}{minminc@google.com}

\icmlkeywords{Machine Learning, ICML}

\vskip 0.3in
]

\newcommand{\jp}[1]{\textcolor{red}{[jp: #1]}}
\newcommand{\mc}[1]{\textcolor{blue}{[mc: #1]}}
\newcommand{\sss}[1]{\textcolor{green}{[sss: #1]}}

\newcommand{\hl}{\mathbf{h}}
\newcommand{\ul}{\mathbf{u}}
\newcommand{\rl}{\mathbf{r}}
\newcommand{\il}{\mathbf{i}}
\newcommand{\xl}{\mathbf{x}}
\newcommand{\zl}{\mathbf{z}}
\newcommand{\bl}{\mathbf{b}}
\newcommand{\wl}{\mathbf{w}}

\newcommand{\Wl}{\mathbf{W}}
\newcommand{\Ul}{\mathbf{U}}



\printAffiliationsAndNotice{\icmlEqualContribution}

\begin{abstract}
Recurrent neural networks have gained widespread use in modeling sequence data across various domains. While many successful recurrent architectures employ a notion of gating, the exact mechanism that enables such remarkable performance is not well understood. We develop a theory for signal propagation in recurrent networks after random initialization using a combination of mean field theory and random matrix theory. To simplify our discussion, we introduce a new RNN cell with a simple gating mechanism that we call the minimalRNN and compare it with vanilla RNNs. Our theory allows us to define a maximum timescale over which RNNs can remember an input. We show that this theory predicts trainability for both recurrent architectures. We show that gated recurrent networks feature a much broader, more robust, trainable region than vanilla RNNs, which corroborates recent experimental findings. Finally, we develop a closed-form critical initialization scheme that achieves dynamical isometry in both vanilla RNNs and minimalRNNs. We show that this results in significantly improved training dynamics. Finally, we demonstrate that the minimalRNN achieves comparable performance to its more complex counterparts, such as LSTMs or GRUs, on a language modeling task. 
\end{abstract}

\section{Introduction}
\label{sec:intro}

Recurrent Neural Networks (RNNs)~\cite{rumelhart1986learning,elman1990finding} have found widespread use across a variety of domains from language modeling~\cite{mikolov2010recurrent,kiros2015skip,jozefowicz2016exploring} and machine translation~\cite{bahdanau2014neural} to speech recognition~\cite{graves2013speech} and recommendation systems~\cite{hidasi2015session,wu2017recurrent}. However, RNNs as originally proposed are difficult to train and are rarely used in practice. Instead, variants of RNNs - such as Long Short-Term Memory (LSTM) networks~\cite{hochreiter1997long} and Gated Recurrent Units (GRU)~\cite{chung2014empirical} - that feature various forms of ``gating'' perform significantly better than their vanilla counterparts. Often, these models must be paired with techniques such as normalization layers~\cite{ioffe2015batch,ba2016layer} and gradient clipping~\cite{pascanu2013difficulty} to achieve good performance.

A rigorous explanation for the remarkable success of gated recurrent networks remains illusive~\cite{jozefowicz2015empirical,greff2017lstm}. Recent work~\cite{collins2016capacity} provides empirical evidence that the benefits of gating are mostly rooted in improved trainability rather than increased capacity or expressivity. The problem of disentangling trainability from expressivity is widespread in machine learning since state-of-the-art architectures are nearly always the result of sparse searches in high dimensional spaces of hyperparameters. As a result, we often mistake trainability for expressivity. Seminal early work~\cite{glorot2010understanding, Bertschinger2005} showed that a major hindrance to trainability was the vanishing and exploding of gradients. 

Recently, progress has been made in the feed-forward setting~\cite{schoenholz2016, pennington2017resurrecting, yang2017mean} by developing a theory of both the forward-propagation of signal and the backward-propagation of gradients. This theory is based on studying neural networks whose weights and biases are randomly distributed. This is equivalent to studying the behavior of neural networks after random initialization or, equivalently, to studying the prior over functions induced by a particular choice of hyperparameters~\cite{lee2017deep}. It was shown that randomly initialized neural networks are trainable if three conditions are satisfied: (1) the size of the output of the network is finite for finite inputs, (2) the output of the network is sensitive to changes in the input, and (3) gradients neither explode nor vanish. Moreover, neural networks achieving dynamical isometry, i.e. having input-output Jacobian matrices that are well-conditioned,  were shown to train orders of magnitude faster than networks that do not.


In this work, we combine mean field theory and random matrix theory to extend these results to the recurrent setting. We will be particularly focused on understanding the role that gating plays in trainability. As we will see, there are a number of subtleties that must be addressed for (gated) recurrent networks that were not present in the feed-forward setting. To clarify the discussion, we will therefore contrast vanilla RNNs with a gated RNN cell, that we call the minimalRNN, which is significantly simpler than LSTMs and GRUs but implements a similar form of gating. We expect the framework introduced here to be applicable to more complicated gated architectures. 

The first main contribution of this paper is the development of a mean field theory for forward propagation of signal through vanilla RNNs and minimalRNNs. 
In doing so, we identify a theory of the maximum timescale over which signal can propagate in each case. 
Next, we produce a random matrix theory for the end-to-end Jacobian of the minimalRNN. 
As in the feed-forward setting, we establish that the duality between the forward propagation of signal and the backward propagation of gradients persists in the recurrent setting. We then show that our theory 
is indeed predictive of trainability in recurrent neural networks by comparing the maximum trainable number of steps of RNNs with the timescale predicted by the theory. Overall, we find remarkable alignment between theory and practice. 
Additionally, we develop a closed-form initialization procedure for both networks and show that on a variety of tasks RNNs initialized to be dynamically isometric are significantly easier to train than those lacking this property. 


Corroborating the experimental findings of~\citet{collins2016capacity}, we show that both signal propagation and dynamical isometry in vanilla RNNs is far more precarious than in the case of the minimalRNN. Indeed the vanilla RNN achieves dynamical isometry only if the network is initialized with orthogonal weights at the boundary between order-and-chaos, a one-dimensional line in parameter space. Owing to its gating mechanism, the minimalRNN on the other hand enjoys a robust multi-dimensional subspace of good initializations which all enable dynamical isometry. Based on these insights, we conjecture that more complex gated recurrent neural networks also benefit from the similar effects.

\section{Related Work}
\label{sec:related}

Identity and Orthogonal initialization schemes have been identified as a promising approach to improve trainability of deep neural networks~\cite{le2015simple,mishkin2015all}. Additionally, \citet{arjovsky2016unitary,hyland2017learning,xie2017all} advocate going beyond initialization to constrain the transition matrix to be orthogonal throughout the entire learning process either through re-parametrisation or by constraining the optimization to the Stiefel manifold~\cite{wisdom2016full}. However, as was pointed out in \citet{vorontsov2017orthogonality}, strictly enforcing orthogonality during training may hinder training speed and generalization performance.  While these contributions are similar to our own, in the sense that they attempt to construct networks that feature dynamical isometry, it is worth noting that orthogonal weight matrices do not guarantee dynamical isometry. This is due to the nonlinear nature of deep neural networks as shown in~\citet{pennington2017resurrecting}. In this paper we continue this trend and show that orthogonality has little impact on the conditioning of the Jacobian (and so trainability) in gated RNNs. 

The notion of ``edge of chaos'' initialization has been explored previously especially in the case of recurrent neural networks. \citet{Bertschinger2005,glorot2010understanding} propose edge-of-chaos initialization schemes that they show leads to improved performance. Additionally, architectural innovations such as batch normalization~\citep{ioffe2015}, orthogonal matrix initialization~\citep{saxe2013exact}, random walk initialization~\citep{sussillo2014}, composition kernels~\citep{daniely2016}, or residual network architectures~\citep{he2015} all share a common goal of stabilizing gradients and improving training dynamics. 

There is a long history of applying mean field-like approaches to understand the behavior of neural networks. Indeed several pieces of seminal work used statistical physics~\citep{derrida1986, sompolinsky1988} and Gaussian Processes~\citep{neal2012} to show that neural networks exhibit remarkable regularity as the width of the network gets large. Mean field theory also has long been used to study Boltzmann machines~\citep{ackley1985} and sigmoid belief networks~\citep{saul1996}. More recently, there has been a revitalization of mean field theory to explore questions of trainability and expressivity in fully-connected networks and residual networks~\citep{poole2016, schoenholz2016,yang2017mean,schoenholz2017correspondence,Karakida2018,hayou2018selection,hanin2018start, yang2018deep}. Our approach will closely follow these later contributions and extend many of their techniques to the case of recurrent networks with gating. Beyond mean field theory, there have been several attempts in understanding signal propagation in RNNs, e.g., using Ger\u{s}gorin circle theorem~\cite{zilly2016recurrent} or time invariance~\cite{tallec2018can}. 



\section{Theory and Critical Initialization}



We begin by developing a mean field theory for vanilla RNNs and discuss the notion of dynamical isometry. Afterwards, we move on to a simple gated architecture to explain the role of gating in facilitating signal propagation in RNNs. 

\subsection{Vanilla RNN}
\label{vanilla}


Vanilla RNNs are described by the recurrence relation,
\begin{align}\label{eq:vanilla_rnn_recurrence}
   \bm e^t = \bm W \bm h^{t-1} + \bm V \bm x^t + \bm b \quad \quad \quad \bm h^t  =  \phi(\bm e^{t}).  
\end{align}
Here $\bm x^t\in \mathbb R^M$ is the input, $\bm e^t\in\mathbb R^N$ is the pre-activation, and $\bm h^t\in\mathbb R^N$ is the hidden state after applying an arbitrary activation function $\phi:\mathbb R\to\mathbb R$. For the purposes of this discussion we set $\phi=\tanh$. Furthermore, $\bm W\in\mathbb R^{N\times N}$ and $\bm V\in\mathbb R^{N\times M}$ are weight matrices that multiply the hidden state and inputs respectively and $\bm b\in\mathbb R^N$ is a bias.

Next, we apply mean-field theory to vanilla RNNs following a similar strategy introduced in~\cite{poole2016,schoenholz2016}. At the level of mean-field theory, vanilla RNNs will prove to be intimately related to feed-forward networks and so this discussion proceeds analogously. For a more detailed discussion, see these earlier studies. 


Consider two sequences of inputs $\{\bm x^t_1\}$ and $\{\bm x^t_2\}$, described by the covariance matrix $\bm R^t \in \mathbb R^{2\times 2}$ with $\bm R_{ab}^t = \frac{1}{M}\mathbb{E}[\bm x_a \cdot \bm x_b],\; a, b \in \{1, 2\}$. 
To simplify notation, we assume the input sequences have been standardized so that $R_{11}^t = R_{22}^t = R$ independent of time. 
This allows us to write $\bm R^t = R\bm \Sigma^t$, where $\bm \Sigma^t$ is a matrix whose diagonal terms are 1 and whose off-diagonal terms are the cosine similarity between the inputs at time $t$. 
These sequences are then passed into two identical copies of an RNN to produce two corresponding pre-activation sequences $\{\bm e^t_1\}$ and $\{\bm e^t_2\}$. As in~\citet{poole2016} we let the weights and biases be Gaussian distributed so that $W_{ij}\sim\mathcal N(0,\sigma_w^2/N)$, $V_{ij}\sim\mathcal N(0, \sigma_v^2/M)$, and $b_i\sim\mathcal N(\mu_b,  \sigma_b^2)$\footnote{in practice we will set $\mu_b = 0$ for vanillaRNN.}, and we consider the wide network limit, $N\to\infty$. As in the fully-connected setting, we would like to invoke the Central Limit Theorem (CLT) to conclude that the pre-activations of hidden states are jointly Gaussian distributed. Unfortunately, the CLT is violated in the recurrent setting as $\bm h^{t-1}$ is correlated with $\bm W$ due to weight sharing between steps of the RNN. 

To make progress, we proceed by developing the theory of signal propagation for RNNs with \emph{untied} weights. This allows for several simplifications, including the application of the CLT to conclude that $e^t_{ia}$ are jointly Gaussian distributed, $$[e^t_{i1}, e^t_{j2}]^T\sim\mathcal N(\mu_b\bm 1, \bm q^t \delta_{ij}), \quad i, j \in \{1, \cdots, N\}$$
where the covariance matrix $\bm q^t \in \mathbb R^{2\times 2}$ is independent of neuron index, $i$. We explore the ramifications of this approximation by comparing simulations of RNNs with tied and untied weights. Overall, we will see that while ignoring weight tying leads to quantitative differences between theory and experiment, it does not change the qualitative picture that emerges. See figs.~\ref{fig:mean_field_minimal} and~\ref{fig:depthscale} for verification.


With this approximation in mind, we will now quantify how the pre-activation hidden states $\{\bm e^t_1\}$ and $\{\bm e^t_2\}$ evolve by deriving the recurrence relation of the covariance matrix $\bm q^t$ from the recurrence on $\bm e^t$ in eq.~\eqref{eq:vanilla_rnn_recurrence}. Using identical arguments to~\citet{poole2016} one can show that,
\begin{align}
\bm q^t = \sigma_w^2\int D_{\bm q^{t-1}}\bm z\;\phi(\bm z)\phi(\bm z)^\top + \sigma_v^2 R \bm \Sigma^t +\sigma_b^2\bm I. \label{eq:mf_recurrence_vanilla}
\end{align}
where $\bm z = [z_1, z_2]^\top,$
and 
\begin{equation}~\label{eq:gaussian_measure}
\int D_{\bm q}\bm z = \frac1{2\pi\sqrt{|\bm q|}}\int d\bm z e^{-(\bm z - \mu_b\bm 1)^T\bm q^{-1}(\bm z - \mu_b\bm 1)}
\end{equation}
is a Gaussian measure with covariance matrix $\bm q$. By symmetry, our normalization allows us to define $q_{11}^t = q_{22}^t = q^t$ to be the magnitude of the pre-activation hidden state and $c^t = q_{12}^t/q^t$ to be the cosine similarity between the hidden states. We will be particularly concerned with understanding the dynamics of the cosine similarity, $c^t$.

In feed-forward networks, the inputs dictate the initial value of the cosine similarity, $c^0$ and then the evolution of $c^t$ is determined solely by the network architecture. By contrast in recurrent networks, inputs perturb $c^t$ at each timestep. Analyzing the dynamics of $c^t$ for arbitrary $\bm \Sigma^t$ is therefore challenging, however significant insight can be gained by studying the off-diagonal entries of eq.~\eqref{eq:mf_recurrence_vanilla} for $\bm \Sigma^t = \bm \Sigma$ independent of time. In the case of time-independent $\bm \Sigma^t$, as $t\to\infty$ both $q^t\to q^\ast$ and $c^t\to c^\ast$ where $q^\ast$ and $c^\ast$ are fixed points of the variance of the pre-activation hidden state and the cosine-similarity between pre-activation hidden states respectively. As was discussed previously~\cite{poole2016,schoenholz2016}, the dynamics of $q^t$ are generally uninteresting provided $q^\ast$ is finite. We therefore choose to normalize the hidden state such that $q^0 = q^\ast$ which implies that $q^t = q^\ast$ independent of time.  

In this setting it was shown in~\citet{schoenholz2016} that in the vicinity of a fixed point, the off-diagonal term in eq.~\eqref{eq:mf_recurrence_vanilla} can be expanded to lowest order in $\epsilon^t = c^\ast - c^t$ to give the linearized dynamics, $\epsilon^{t} = \chi_{c^\ast}\epsilon^{t-1}$ where
\begin{equation}\label{eq:chi_cstar_vanilla}
    \chi_{c^\ast} = \sigma_w^2\int D_{\bm q^*}\bm z\phi'(z_1)\phi'(z_2).
\end{equation}
These dynamics have the solution $\epsilon^t = \chi_{c^*}^{t-t_0}\epsilon^{t_0}$ where $t_0$ is the time when $c^t$ is close enough to $c^*$ for the linear approximation to be valid. If $\chi_{c^*} < 1$ it follows that $c^t$ approaches $c^\ast$ exponentially quickly over a timescale $\tau = -1/\log\chi_{c^\ast}$ and $c^\ast$ is called a stable fixed point. When $c^t$ gets too close to $c^\ast$ to be distinguished from it to within numerical precision, information about the initial inputs has been lost. Thus, $\tau$ sets the maximum timescale over which we expect the RNN to be able to remember information. If $\chi_{c^\ast} > 1$ then $c^t$ gets exponentially farther from $c^\ast$ over time and $c^\ast$ is an unstable fixed point. In this case, for the activation function considered here, another fixed point that is stable will emerge. Note that $\chi_{c^\ast}$ is independent of $\bm\Sigma$ and so the dynamics of $c^t$ near $c^\ast$ do not depend on $\bm\Sigma$. 

In vanilla fully-connected networks $c^\ast = 1$ is always a fixed point of $c^t$, but it is not always stable. Indeed, it was shown that these networks exhibit a phase transition where $c^\ast = 1$ goes from being a stable fixed point to an unstable one as a function of the network's hyperparameters. This is known as the order-to-chaos transition and it occurs exactly when $\chi_1 = 1$. Since $\tau = -1/\log(\chi_1)$,  signal can propagate infinitely far at the boundary between order and chaos.  Comparing the diagonal and off-diagonal entries of eq.~\eqref{eq:mf_recurrence_vanilla}, 
we see that in recurrent networks, $c^\ast = 1$ is a fixed point only when $\Sigma_{12} = 1$, and in this case the discussion is identical to the feed-forward setting. When $\Sigma_{12} < 1$, it is easy to see that $c^\ast < 1$ since if $c^t = 1$ at some time $t$ then $c^{t+1} = 1 - \sigma_v^2R(1-\Sigma_{12})/q^\ast < 1$. We see that in recurrent networks noise from the inputs destroys the ordered phase and there is no ordered-to-chaos critical point. As a result we should expect the maximum timescale over which memory may be stored in vanilla RNNs to be fundamentally limited by noise from the inputs. 

The end-to-end Jacobian  of a vanilla RNN with untied weights is in fact formally identical to the input-output Jacobian of a feedforward network, and thus the results from~\cite{pennington2017resurrecting} regarding conditions for dynamical isometry apply directly. In particular, dynamical isometry is achieved with orthogonal state-to-state transition matrices $\bm W$, $\tanh$ non-linearities, and small values of $q^\ast$. 
Perhaps surprisingly, these conclusions continue to be valid if the assumption of untied weights is relaxed. To understand why this is the case, consider the example of a linear network. For untied weights, the end-to-end Jacobian is given by $\hat{\bm J} = \prod_{t=1}^T \bm W_t$, while for tied weights the Jacobian is given by ${\bm J} = {\bm W}^T$. It turns out that as $N\to\infty$ there is sufficient self-averaging to overcome the dependencies induced by weight tying and the asymptotic singular value distributions of $\hat{\bm J}$ and $\bm J$ are actually identical \cite{haagerup2000brown}.



\subsection{MinimalRNN}

\subsubsection{Mean-Field Theory}



To study the role of gating, we introduce the minimalRNN which is simpler than other gated RNN architectures but nonetheless features the same gating mechanism. A sequence of inputs $\bm{x}^t\in\mathbb R^M$, is first mapped to the hidden space through
$\bm{\tilde{x}}^t = \Phi(\bm{x}^t)$\footnote{$\Phi(\cdot)$ here can be any highly flexible functions such as a feed-forward network. In our experiments, we take $\Phi(\cdot)$ to be a fully connected layer with $\tanh$ activation, that is, $\Phi(\bm{x}^t) = \tanh(\bm W_x\bm x^t)$.}. From here on, we refer to $\bm{\tilde{x}}^t \in \mathbb R^N$ as the inputs to minimalRNN.
The minimalRNN is then described by the recurrence relation,
\begin{align}\label{eq:minimal_rnn_h}
\bm{e}^t  & = \bm{W} \bm{h}^{t-1} + \bm{V} \bm{\tilde x}^t + \bm{b} \quad \quad \quad \bm{u}^t = \sigma(\bm e^t)\\
\bm{h}^t & = \bm{u}^t \odot \bm{h}^{t-1} + (\bm{1}-\bm{u}^t) \odot \bm{\tilde x}^t \nonumber
\end{align}
where 
$\bm e^t \in \mathbb R^N$ is the pre-activation to the gating function,  $\bm{u}^t \in \mathbb R^N$ the update gate and $\bm h^t \in \mathbb{R}^N$ the hidden state. The minimalRNN retains the most essential gate in LSTMs~\cite{jozefowicz2015empirical,greff2017lstm} and achieves competitive performance. The simplified update of this cell on the other hand, enables us to pinpoint the role of gating in a more controlled setting. 

As in the previous case we consider two sequences of inputs to the network, $\{\bm{\tilde x}_1^t\}$ and $\{\bm{\tilde x}_2^t\}$. We take $W_{ij}\sim\mathcal N(0, \sigma_w^2/N)$, $V_{ij}\sim\mathcal N(0, \sigma_v^2/N)$ and $b_i\sim\mathcal N(\mu_b,\sigma_b^2)$. 
By analogy to the vanilla case, we can make the mean field approximation that the $e^t_{ia}$ are jointly Gaussian distributed with covariance matrix $\bm q^t\delta_{ij} \in \mathbb{R}^{2\times2}$. Here,
\begin{equation}~\label{eq:minimal_rnn_q_expectation}
\bm q^t = \sigma_w^2 \bm Q^{t-1} + \sigma_v^2 \bm R^t + \sigma_b^2\bm I
\end{equation}
where we have defined $\bm Q^t$ as the second-moment matrix with $Q_{ab}^t =  \mathbb  E[h_{ia}^t h_{ib}^t]$.~\footnote{$\bm h^t$ will be centered under mean field approximation if $\bm h^0$ is initialized with mean zero.}
As in the vanilla case, $\bm R^t$ is the covariance between inputs so that $R^t_{ab} = \frac{1}{N}\mathbb E[\bm{\tilde x}_a \cdot \bm{\tilde x}_{b}]$.  

We note that $\bm R^t$ is fixed by the input, but it remains for us to work out $\bm Q^t$. We find that (see SI section~\ref{supp:Q_recurrence}),
\begin{align}
    \bm Q^t = \bm Q^{t-1}&\int D_{\bm q^t}\bm z\;\sigma(\bm z)\sigma(\bm z)^\top ~\label{eq:minimal_rnn_expectation}\\
     & + \bm R^t\int D_{\bm q^t}\bm z\;(1-\sigma(\bm z))(1-\sigma(\bm z))^\top\nonumber
\end{align}
Here we assume that the expectation factorizes so that $\bm h^{t-1}$ and $\bm u^t$ are approximately independent. We believe this approximation becomes exact in the $N\to\infty$ limit.

We choose to normalize the data in a similar manner to the vanilla case so that $R_{11}^t = R_{22}^t = R$ independent of time. An immediate consequence of this normalization is that $Q_{11}^t = Q_{22}^t = Q^t$ and $q_{11}^t = q_{22}^t = q^t$. 
We then write $C^t = Q_{12}^t / Q^t$ and $c^t = q_{12}^t / q^t$ as the cosine similarities between the hidden states and the pre-activations respectively. 
With this normalization, we can work out the mean-field recurrence relation characterizing the covariance matrix for the minimalRNN. This analysis can be done by  deriving the recurrence relation for either $\bm Q^t$ or $\bm q^t$. We will choose to study the dynamics of $\bm q^t$, however the two are trivially related by eq.~\eqref{eq:minimal_rnn_q_expectation}. In SI section~\ref{supp:Q_fixed_point}, we analyze the dynamics of the diagonal term in the recurrence relation and prove that there is always a fixed point at some $q^\ast$. In SI section~\ref{supp:Q_dynamics}, we compute the depth scale over which $q^t$ approaches $q^\ast$. However, as in the case of the vanillaRNN, the dynamics of $q^\ast$ are generally uninteresting.

We now turn our attention to the dynamics of the cosine similarity between the pre-activations, $c^t$. As in the case of vanilla RNNs, we note that $q^t$ approaches $q^\ast$ quickly relative to the dynamics of $c^t$. We therefore choose to normalize the hidden state of the RNN so that $Q^0 = Q^\ast$ in which case both $Q^t=Q^\ast$ and $q^t=q^\ast$ independent of time. From eq.~\eqref{eq:minimal_rnn_q_expectation} and ~\eqref{eq:minimal_rnn_expectation} it follows that the cosine similarity of the pre-activation evolves as,
\begin{align}\label{eq:minimal_rnn_off_diagonal_raw}
c^t = &\left[c^{t-1}  + (\sigma_w^2 - \sigma_v^2)\rho^{t-1}\right] \int D_{\bm q^{t-1}}\bm z\;\sigma(z_1)\sigma(z_2) \nonumber\\
 &- 2\sigma_w^2\rho^{t-1} \int D_{\bm q^{t-1}}\bm z\; \sigma(z_1) + \sigma_w^2 \rho^{t-1} +  \sigma_v^2 \rho^t
\end{align}
where we have defined $\rho^t = R \Sigma_{12}^t / q^\ast$. As in the case of the vanilla RNN, we can study the behavior of $c^t$ in the vicinity of a fixed point, $c^*$. By expanding eq.~\eqref{eq:minimal_rnn_off_diagonal_raw} to lowest order in $\epsilon^t = c^* - c^t$ we arrive at a linearized recurrence relation that has an exponential solution $\epsilon^{t+1} = \chi_{c^*}\epsilon^t$ where here,
\begin{align}\label{eq:minimal_chi_c}
    \chi_{c^*} &= \int D_{\bm q^*}\bm z\sigma(z_1)\sigma(z_2)\\ 
    &+ \left(q^\ast c^\ast + (\sigma_w^2 - \sigma_v^2)R\Sigma_{12}\right)\int D_{\bm q^*}\bm z\sigma'(z_1)\sigma'(z_2).\nonumber
\end{align}
The discussion above in the vanilla case carries over directly to the minimalRNN with the appropriate replacement of $\chi_{c^*}$. Unlike in the case of the vanilla RNN, here we see that $\chi_{c^*}$ itself depends on $\Sigma_{12}$. 

Again $c^* = 1$ is a fixed point of the dynamics only when $\Sigma_{12} = 1$. In this case, the minimalRNN experiences an order-to-chaos phase transition when $\chi_1 = 1$ at which point the maximum timescale over which signal can propagate goes to infinity. Similar to the vanilla RNN, when $\Sigma_{12} < 1$, we expect that the phase transition will be destroyed and the maximum duration of signal propagation will be severely limited. However, in a significant departure from the vanilla case, when $\mu_b\to\infty$ we notice that $\sigma(z + \mu_b) \to 1$, and $\sigma'(z + \mu_b)\to 0$ for all $z$. Considering eq.~\eqref{eq:minimal_chi_c} we notice that in this regime $\chi_{c^*}\to 1$ independent of $\Sigma_{12}$. In other words, gating allows for arbitrarily long term signal propagation in recurrent neural networks independent of $\Sigma_{12}$.

\begin{figure*}[t]
    \centering
    \includegraphics[width=1.0\textwidth]{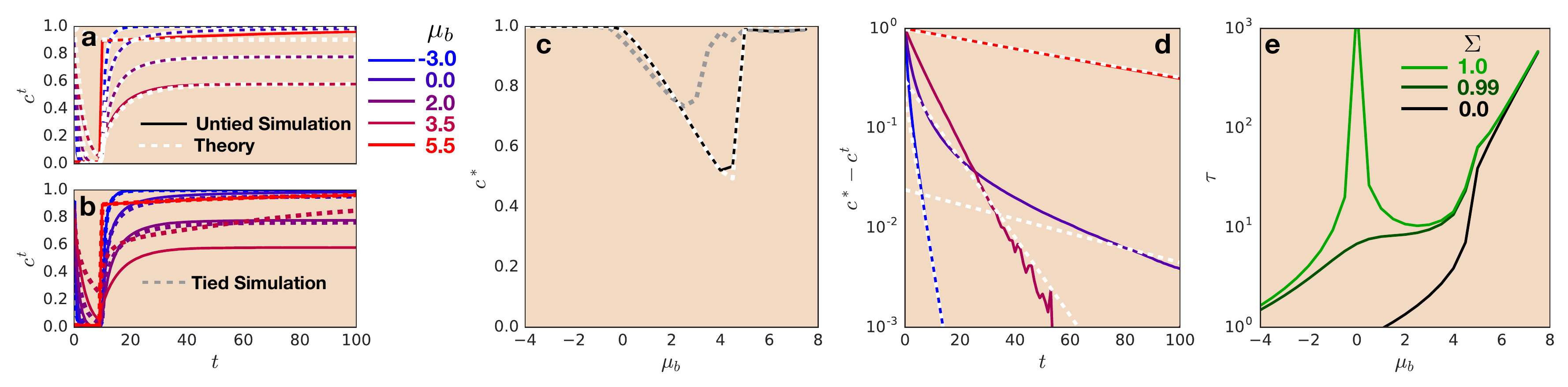}
    \vspace{-0.2in}
    \caption{\textbf{Numerical verification of mean field results on the minimalRNN.} All experiments were done with fixed $\sigma_w = 6.88$, $\sigma_v = 1.39$, $R = 0.46$, and $\mu_b\in[-4,8]$. These hyperparameters are chosen so that the minimalRNN has an order-to-chaos critical point at $\mu_b = 0$. MC simulations were averaged over 1000 minimalRNNs with hidden dimension 8192. (a) Cosine similarity for theory (white dashed) compared with MC simulations of a minimalRNN with untied weights (solid). (b) Cosine similarity for tied weights (dashed) compared with untied weights (solid). (c) Comparison of $c^*$ for theory (white dashed), MC with untied weights (solid black), tied weights (dashed grey). (d) Comparison of linearized dynamics. Dashed lines show $A e^{-t/\tau}$ and solid lines show the result of simulations with untied weights. (e) Comparison of the timescale for signal propagation for different values of $\Sigma_{12}$.}
    \label{fig:mean_field_minimal}
\end{figure*}

We explore agreement between our theory and MC simulations of the minimalRNN in fig.~\ref{fig:mean_field_minimal}. In this set of experiments, we consider inputs such that $\Sigma_{12}^t = 0$ for $t < 10$ and $\Sigma_{12}^t = 1$ for $t\geq 10$. Fig.~\ref{fig:mean_field_minimal} (a,c,d) show excellent quantitative agreement between our theory and MC simulations. In fig.~\ref{fig:mean_field_minimal} (a,b) we compare the MC simulations of the minimalRNN with and without weight tying. While we observe that for many choices of hyperparameters the untied weight approximation is quite good (particularly when $c^\ast\approx 1$), deeper into the chaotic phase the quantitative agreement between breaks down. Nonetheless, we observe that the untied approximation describes the qualitative behavior of the real minimalRNN overall. In fig.~\ref{fig:mean_field_minimal} (e) we plot the timescale for signal propagation for $\Sigma_{12} = 1, 0.99,$ and $0$ for the minimalRNN with identical choices of hyperparameters. We see that while $\tau\to\infty$ as $\mu_b$ gets large independent of $\Sigma_{12}$, a critical point at $\mu_b = 0$ is only observed when $\Sigma_{12} = 1$. 

\subsubsection{Dynamical Isometry}

In the previous subsection, we derived a quantity $\chi_1$ that defines the boundary between the ordered and the chaotic phases of forward propagation. Here we show that it also defines the boundary between exploding and vanishing gradients. To see this, consider the Jacobian of the state-to-state transition operator,
\begin{equation}
    \bm{J}_t  = \frac{\partial \bm{h}^{t+1}}{\partial \bm{h}^t} = \bm{D}_{\bm{u}^t} + \bm{D}_{\sigma'(\bm{e}^t)\odot(\bm{h}^{t-1}-\bm{z}^t)} \bm{W}\,,
\end{equation}
where $\bm{D}_{\bm{x}}$ denotes a diagonal matrix with $\bm{x}$ along its diagonal. We can compute the expected norm-squared of back-propagated error signals, which measures the growth or shrinkage of gradients. It is equal to the mean-squared singular value of the Jacobian~\cite{poole2016,schoenholz2016} or the first moment of $\bm{J}_t\bm{J}_t^T$,
\begin{equation}
\frac{1}{N}\mathbb E[\tr(\bm{J}_t\bm{J}_t^T)] = \mathbb E[(u^t_1)^2] + \sigma_w^2 \mathbb E[\sigma'(e^t_1)^2(h^{t-1}_1-z_1^t)^2],  
\end{equation}
where we have used the fact that the elements of $\bm u^t$, $\bm h^t$ and $\bm z^t$ are i.i.d. Since we assume convergence to the fixed point, these distributions are independent of $t$ and it is easy to see that $\frac{1}{N}\mathbb E[\tr(\bm{J}_t\bm{J}_t^T)] = \chi_1$. The variance of back-propagated error signals through $T$ time steps is therefore $\frac{1}{N}\mathbb E[\tr({\bm J}{\bm J}^T)] = \chi_1^T$. As such, the constraint $\chi_1 = 1$ defines the boundary between phases of exponentially exploding and exponentially vanishing gradient norm (variance). Note that unlike in the case of forward signal propagation, in the case of backpropagation this is independent of $\bm\Sigma$.

As argued in~\cite{pennington2017resurrecting,PenningtonSG18}, controlling the variance of back-propagated gradients is necessary but not sufficient to guarantee trainability, especially for very deep networks. Beyond the first moment, the entire distribution of eigenvalues of ${\bm J}{\bm J}^T$ (or of singular values of ${\bm J}$) is relevant. Indeed, it was found in~\cite{pennington2017resurrecting,PenningtonSG18} that enabling dynamical isometry, namely the condition that all singular values of ${\bm J}$ are close to unity, can drastically improve training speed for very deep feed-forward networks.

Following~\cite{pennington2017resurrecting,PenningtonSG18}, we use tools from free probability theory to compute the variance $\sigma^2_{{\bm J}{\bm J}^T}$ of the limiting spectral density of ${\bm J}{\bm J}^T$; however, unlike previous work, in our case the relevant matrices are not symmetric and therefore we must invoke tools from non-Hermitian free probability, see~\cite{cakmak2012non} for a review. As in previous section, we make the simplifying assumption that the weights are untied, relying on the same motivations given in section~\ref{vanilla}. Using these tools, an un-illuminating calculation reveals that,
\begin{equation}
\label{eq:eigen_variance}
    \sigma^2_{{\bm J}{\bm J}^T} = \chi_1^{2T} \left(1 + T \frac{2(\mu_1-s_1) \mu_2+\sigma_1^2 + \sigma_2^2}{\chi_1^2} \right)\,,
\end{equation}
where,
\begin{align}
\label{eq:minimal_chi1}
\chi_1 & = \mu_1 + \mu_2 \\
\mu_1 &= \int \mathcal{D}z\;\sigma^2(\sqrt{q^*} z + \mu_b)\nonumber \\
\sigma_1^2 &= -\mu_1^2 + \int \mathcal{D}z\; \sigma^4(\sqrt{q^*} z + \mu_b)\nonumber\\
\mu_2 &= \sigma_{w}^2 (Q^* + R)\int \mathcal{D}z\;[\sigma'(\sqrt{q^*}z+\mu_b)]^2\nonumber\\
\sigma_2^2 &= -\mu_2^2 + \sigma_{w}^4 ((Q^*)^2 + R^2)\int \mathcal{D}z\;[\sigma'(\sqrt{q^*}z+\mu_b)]^4\nonumber
\end{align}
and $s_1$ is the first term in the Taylor expansion of the S-transform of the eigenvalue distribution of $WW^T$~\cite{PenningtonSG18}. For example, for Gaussian matrices, $s_1 = -1$ and for orthogonal matrices $s_1 = 0$.

Some remarks are in order about eq.~(\ref{eq:eigen_variance}). First, we note the duality between the forward and backward signal propagation (eq.~(\ref{eq:minimal_chi_c}) and eq.~(\ref{eq:minimal_chi1})).  For critical initializations, $\chi_1 = 1$, so $\sigma^2_{{\bm J}{\bm J}^T}$ does not grow exponentially, but it still grows linearly with $T$. This situation is entirely analogous to the feed-forward analysis of~\cite{pennington2017resurrecting,PenningtonSG18}. In the case of the vanilla RNN, the coefficient of the linear term is proportion to $q^*$, and can only be reduced by taking the weight and bias variances $(\sigma_w^2, \sigma_b^2) \to (1,0)$. A crucial difference in the minimalRNN is that the coefficient of the linear term can be made arbitrarily small by simply adjusting the bias mean $\mu_b$ to be positive, which will send $\mu_2 \to 0$ and $\mu_1 \to 1$ independent of $\Sigma$. Therefore the conditions for dynamical isometry decouple from the weight and bias variances, implying that trainability can occur for a higher-dimensional, more robust, slice of parameter space. Moreover, the value of $s_1$ has no effect on the capacity of the minimalRNN to achieve dynamical isometry. We believe these are fundamental reasons why gated cells such as the minimalRNN perform well in practice.

Algorithm~\ref{alg:crit_init} describes the procedure to find $\sigma_w^2, \sigma_v^2$ and $\sigma_b^2$ to achieve $\chi_1$ condition for minimalRNN. Given $\sigma_w^2, \sigma_v^2, \sigma_b^2$, we then construct the weight matrices and biases accordingly. $Q^\ast$ is used to initialize the $h^0$ to avoid transient phase. 

\begin{algorithm}
\caption{Critical initialization for minimalRNNs}\label{alg:crit_init}
\begin{algorithmic}[1]
  \REQUIRE $q^\ast,\mu_b,R$
  \STATE $\mathbb{E}[u^2] \gets \int \mathcal{D}z\; \sigma^2(\sqrt{q^*} z + \mu_b) $
  \STATE $\mathbb{E}[(1-u)^2] \gets \int \mathcal{D}z\; \left(1-\sigma(\sqrt{q^*} z + \mu_b)\right)^2 $
  \STATE $Q^\ast \gets R \cdot \mathbb{E}[(1-u)^2] / (1 - \mathbb{E}[u^2])\quad$ (eq.\eqref{eq:minimal_rnn_expectation})
  \STATE $\mathbb{E}[u'^2] \gets \int \mathcal{D}z[ \sigma'(\sqrt{q^*} z + \mu_b)]^2 $
  \STATE $\sigma_w^2 \gets (1-\mathbb{E}[u^2]) /(Q^\ast + R) / \mathbb{E}[u'^2]\quad$ (eq.\eqref{eq:minimal_chi1})
  \STATE $\sigma_b^2 \gets 0$
  \STATE $\sigma_v^2 \gets (q^\ast - Q^\ast\sigma_w^2 - \sigma_b^2)/R$
\end{algorithmic}
\end{algorithm}

\section{Experiments}
\label{exp}

\begin{figure*}[!ht]
\centering
\includegraphics[width=0.9\textwidth]{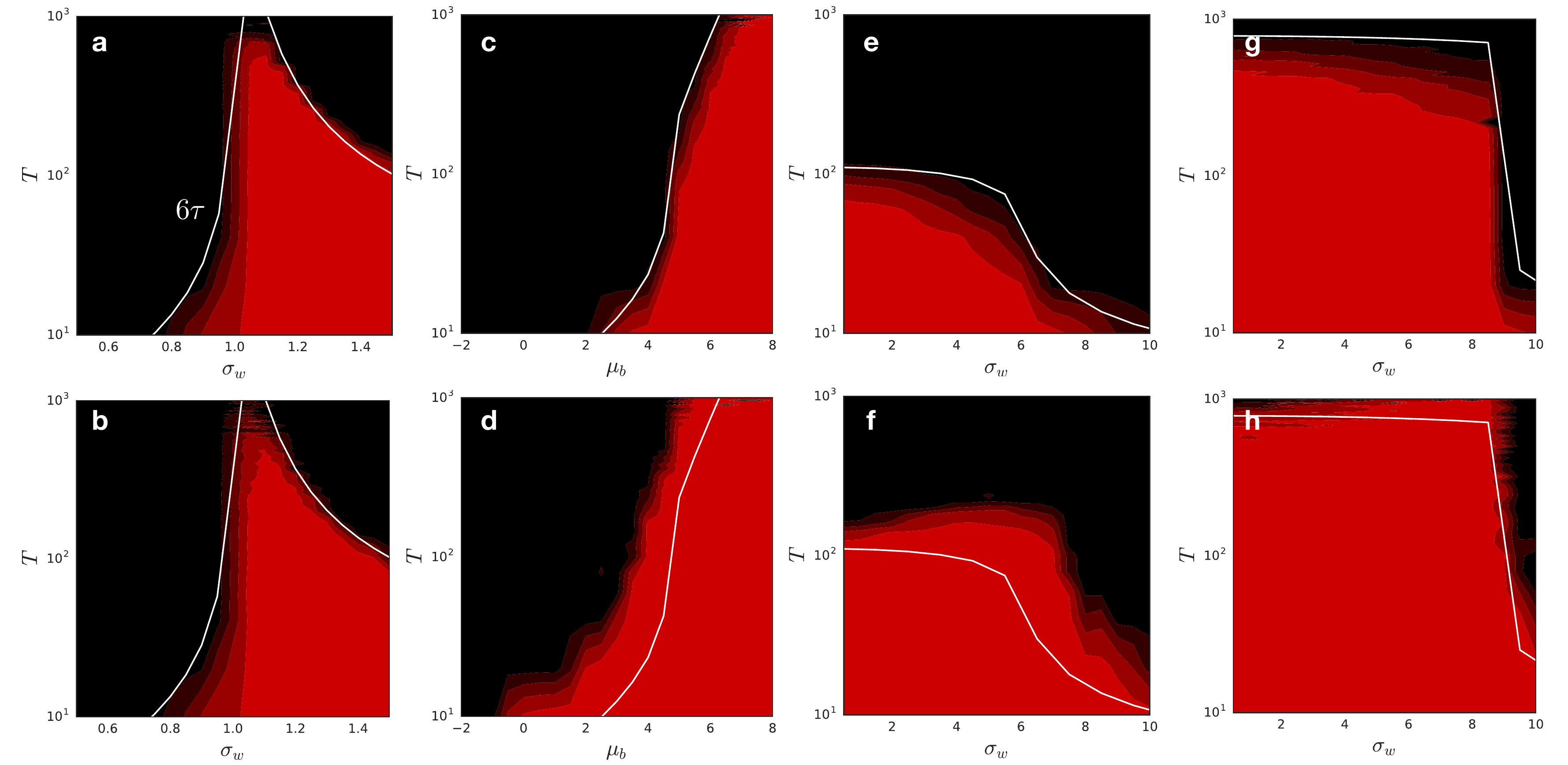} 
  \caption{\textbf{Relationship between theory and trainability.} We plot the training accuracy (higher accuracies in red) overlayed with theoretical timescale, $\tau$ (shown in white). The top row of figures shows the results with untied weights and the bottom row shows the results with weight tying. (a-b) Vanilla RNN with $\sigma_w\in[0.5,1.5]$. (c-d) MinimalRNN with $\mu_b\in[-4,8]$. (e-f) MinimalRNN with $\sigma_w\in[0.5,10]$ and $\mu_b = 4$. (g-h) MinimalRNN with $\sigma_w\in[0.5,10]$ and $\mu_b = 6$.}
  \label{fig:depthscale}
\end{figure*}

Having established a theory for the behavior of random vanilla RNNs and minimalRNNs, we now discuss the connection between our theory and trainability in practice. We begin by corroborating the claim that the maximum timescale over which memory can be stored in a RNN is controlled by the timescale $\tau$ identified in the previous section. We will then investigate the role of dynamical isometry in speeding up learning.

\subsection{Trainability}

\textbf{Dataset.} To verify the results of our theoretical calculation, we consider a task that is reflective of the theory above. To that end, we constructed a sequence dataset for training RNNs from MNIST~\cite{lecun1998gradient}.  Each of the $28\times 28$ digit image is flattened into a vector of $784$ pixels and sent as the first input to a RNN.  We then send $T$ random inputs $\bm x^t\sim\mathcal N(0, \sigma_x^2)$, $0 < t < T$ into the RNN varying $T$ between 10 and 1000 steps. As the only salient information about the digit is in the first layer, the network will need to propagate information through $T$ layers to accurately identify the MNIST digit. The random inputs are drawn independently for each example and so this is a regime where $\Sigma^t = 0$ for all $t > 0$.

We then performed a series of experiments on this task to make connection with our theory. In each case we experimented with both tied and untied weights. The result are shown in fig.~\ref{fig:depthscale}. In the case of untied weights, we observe strong quantitative agreement between our theoretical prediction for $\tau$ and the maximum depth $T$ where the network is still trainable. When the weights of the network are tied, we observe quantitative deviations between our thoery and experiments, but the overall qualitative picture remains. 

We train vanilla RNNs for $10^3$ steps (around 10 epochs) varying $\sigma_w\in[0.5,1.5]$ while fixing $\sigma_v = 0.025$. The results of this experiment are shown in fig.~\ref{fig:depthscale} (a-b). We train minimalRNNs for $10^2$ steps (around 1 epoch) fixing $\sigma_v = 1.39$. We perform three different experiments here: 1) varying $\mu_b\in[-4,8]$ with $\sigma_w=6.88$ shown in fig.~\ref{fig:depthscale} (c-d), 2) varying $\sigma_w\in[0.5,10]$ with $\mu_b = 4$ shown in fig.~\ref{fig:depthscale} (e-f), 3) varying $\sigma_w\in[0.5,10]$ with $\mu_b = 6$ shown in fig.~\ref{fig:depthscale} (g-h). Comparing fig.~\ref{fig:depthscale}(a,b) with fig.~\ref{fig:depthscale}(c,d, g,h), the minimalRNN with large depth $T$ is trainable over a much wider range of hyperparameters than the vanillaRNN despite the fact that the network was trained for an order of magnitude less time. 

\subsection{Critical initialization}


\begin{figure*}[!ht]
    \centering
    \includegraphics[width=0.92\textwidth]{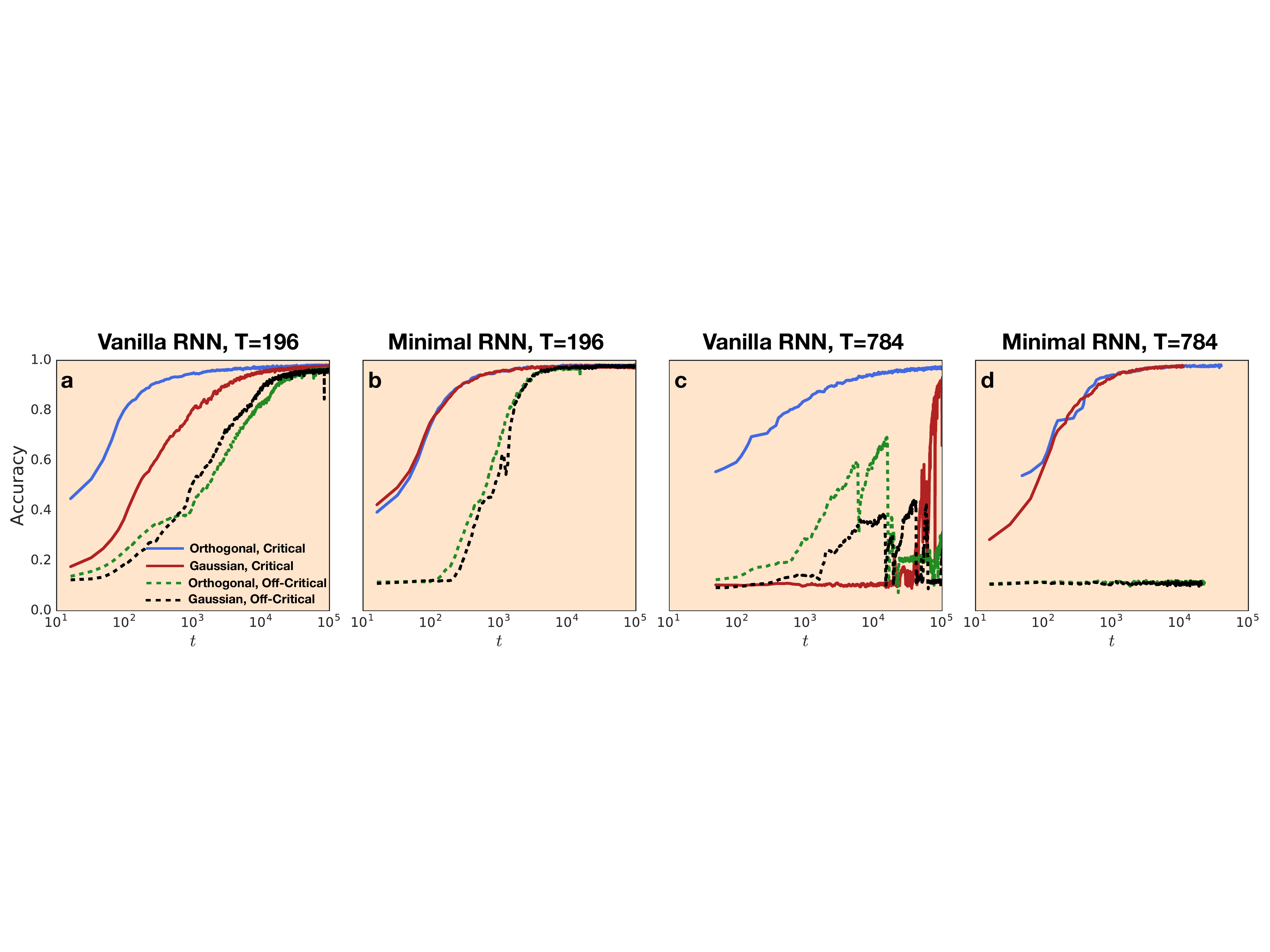} 

  \caption{Learning dynamics, measured by accuracy on the test set, for vanillaRNNs and minimalRNNs trained with depth 196 (a, b) and 784 (c, d) under four different initialization conditions. Drastic difference in terms of convergence speed was observed between critical and off-critical initialization in both models. Well-trained models reach an accuracy of 0.98 on the test set.}
  \label{fig:critical_init}
\end{figure*}

\textbf{Dataset.} To study the impact of critical initialization on training speed, we constructed a more realistic sequence dataset from MNIST. We unroll the pixels into a sequence of $T$ inputs, each containing $784/T$ pixels. We tested $T=196$ and $T=784$ to vary the difficulty of the tasks. 

Note that we are more interested in the training speed of these networks under different initialization conditions than the test accuracy. We compare the convergence speed of vanilla RNN and minimalRNN under four initialization conditions: 1) critical initialization with orthogonal weights (solid blue); 2) critical initialization with Gaussian distributed weights (sold red); 3) off-critical initialization with orthogonal weights (dotted green); 4) off-critical initialization with Gaussian distributed weights (dotted black).




We fix $\sigma_b^2$ to zero in all settings. Under critical initialization,  $\sigma_w^2$ and  $\sigma_v^2$ are carefully chosen to achieve $\chi_1=1$ as defined in eqn.(\ref{eq:chi_cstar_vanilla}) for vanilla RNN and eqn.(\ref{eq:minimal_chi1}) (detailed in algorithm~\ref{alg:crit_init}) for minimalRNN respectively. 
When testing networks off criticality, we employ a common initialization procedure in which, $\sigma_w^2 = 1.0$ and $\sigma_v^2 = 1.0$.  

Figure~\ref{fig:critical_init} summarizes our findings: 
there is a clear difference in training speed between models trained with critical initialization compared with models initialized far from criticality. We observe two orders of magnitude difference in training speed between a critical and off-critical initialization for vanilla RNNs. While a critically initialized model reaches a test accuracy of $90\%$ after 750 optimization steps, the off-critical nework takes over 16,000 updates. A similar trend was observed for the minimalRNN. 
This difference is even more pronounced in the case of the longer sequence with $T=784$. Both vanilla RNNs and minimalRNNs initialized off-criticality failed at task. The well-conditioned minimalRNN trains a factor of three faster than the vanilla RNN. As predicted above, the difference in training speed between orthogonal and Gaussian initialization schemes is significant for vanilla RNNs but is insignificant for the minimalRNN. This is corroborated in fig.~\ref{fig:critical_init} (b,d) where the distribution of the weights has no impact on the training speed.

\section{Language modeling}
\label{ptb}

We compare the minimalRNN against more complex gated RNNs such as LSTM and GRU on the Penn Tree-Bank corpus~\cite{marcus1993building}. Language modeling is a difficult task, and competitive performance is often achieved by more complicated RNN cells. We show that the minimalRNN achieves competitive performance despite its simplicity. 

We follow the precise setup of ~\cite{mikolov2010recurrent,zaremba2014recurrent}, 
 and train RNNs of two sizes: a small configuration with 5M parameters and a medium-sized configuration with 20M parameters~\footnote{The hidden layer size of these networks are adjusted accordingly to reach the target model size.}. We report the perplexity on the validation and test sets. We focus our comparison on single layer RNNs, however we also report perplexities for multi-layer RNNs from the literature for reference. We follow the learning schedule of~\citet{zaremba2014recurrent} and ~\cite{jozefowicz2015empirical}. We review additional hyperparameter ranges in section~\ref{supp:ptb} of the supplementary material.

Table~\ref{tbl:ptb} summarizes our results. We find that single layer RNNs perform on par with their multi-layer counterparts. Despite being a significantly simpler model, the minimalRNN performs comparably to GRUs. Given the closed-form critical initialization developed here that significantly boosts convergence speed, the minimalRNN might be a favorable alternative to GRUs. There is a gap in perplexity between the performance of LSTMs and minimalRNNs. We hypothesize that this is due to the removal of an independent gate on the input. The same strategy is employed in GRUs and may cause a conflict between keeping longer-range memory and updating new information as was originally pointed out by~\citet{hochreiter1997long}. 

\begin{table}
\begin{tabular}{|l|r|r|r|}
\hline
  & 5M-t & 20M-v & 20M-t\\
\hline
 VanillaRNN {\scriptsize{\cite{jozefowicz2015empirical}}}  & 122.8 & 103.0 & 97.7 \\
 GRU\quad\quad\quad {\scriptsize{\cite{jozefowicz2015empirical}}} & 108.2 & 95.5 & 91.7 \\
 LSTM \quad\quad\,{\scriptsize{\cite{jozefowicz2015empirical}}}  & 109.7 &83.3 & 78.8 \\
 \hline
 \hline
 LSTM    & 95.4 &  87.5 & 83.8 \\
  GRU    & 99.5  & 93.9 & 89.8 \\
minimalRNN   & 101.4 & 94.4 & 89.9\\
\hline
\end{tabular}
\caption{Perplexities on the PTB. minimalRNN achieves comparable performance to the more complex gated RNN architectures despite its simplicity.}
\label{tbl:ptb}
\vspace{-0.1in}
\end{table}

\section{Discussion} 

We have developed a theory of signal propagation for random vanilla RNNs and a simple gated RNNs. We demonstrate rigorously that the theory predicts trainability of these networks and gating mechanisms allow for a significantly larger trainable region. We are planning to extend the theory to more complicated RNN cells as well as RNNs with multiple layers.

\section*{Acknowledgements}

We thank Jascha Sohl-Dickstein and Greg Yang for helpful discussions and Ashish Bora for many contributions to early stages of this project.

\bibliography{icml2018_main}

\begin{thebibliography}{54}
\providecommand{\natexlab}[1]{#1}
\providecommand{\url}[1]{\texttt{#1}}
\expandafter\ifx\csname urlstyle\endcsname\relax
  \providecommand{\doi}[1]{doi: #1}\else
  \providecommand{\doi}{doi: \begingroup \urlstyle{rm}\Url}\fi

\bibitem[Ackley et~al.()Ackley, Hinton, and Sejnowski]{ackley1985}
Ackley, David~H., Hinton, Geoffrey~E., and Sejnowski, Terrence~J.
\newblock A learning algorithm for boltzmann machines*.
\newblock \emph{Cognitive Science}, 9\penalty0 (1).
\newblock ISSN 1551-6709.

\bibitem[Arjovsky et~al.(2016)Arjovsky, Shah, and Bengio]{arjovsky2016unitary}
Arjovsky, Martin, Shah, Amar, and Bengio, Yoshua.
\newblock Unitary evolution recurrent neural networks.
\newblock In \emph{International Conference on Machine Learning}, pp.\
  1120--1128, 2016.

\bibitem[Ba et~al.(2016)Ba, Kiros, and Hinton]{ba2016layer}
Ba, Jimmy~Lei, Kiros, Jamie~Ryan, and Hinton, Geoffrey~E.
\newblock Layer normalization.
\newblock \emph{arXiv preprint arXiv:1607.06450}, 2016.

\bibitem[Bahdanau et~al.(2014)Bahdanau, Cho, and Bengio]{bahdanau2014neural}
Bahdanau, Dzmitry, Cho, Kyunghyun, and Bengio, Yoshua.
\newblock Neural machine translation by jointly learning to align and
  translate.
\newblock \emph{arXiv preprint arXiv:1409.0473}, 2014.

\bibitem[Bertschinger et~al.()Bertschinger, Natschl\"{a}ger, and
  Legenstein]{Bertschinger2005}
Bertschinger, Nils, Natschl\"{a}ger, Thomas, and Legenstein, Robert~A.
\newblock At the edge of chaos: Real-time computations and self-organized
  criticality in recurrent neural networks.

\bibitem[Cakmak(2012)]{cakmak2012non}
Cakmak, Burak.
\newblock Non-hermitian random matrix theory for mimo channels.
\newblock Master's thesis, Institutt for elektronikk og telekommunikasjon,
  2012.

\bibitem[Chung et~al.(2014)Chung, Gulcehre, Cho, and
  Bengio]{chung2014empirical}
Chung, Junyoung, Gulcehre, Caglar, Cho, KyungHyun, and Bengio, Yoshua.
\newblock Empirical evaluation of gated recurrent neural networks on sequence
  modeling.
\newblock \emph{arXiv preprint arXiv:1412.3555}, 2014.

\bibitem[Collins et~al.(2016)Collins, Sohl-Dickstein, and
  Sussillo]{collins2016capacity}
Collins, Jasmine, Sohl-Dickstein, Jascha, and Sussillo, David.
\newblock Capacity and trainability in recurrent neural networks.
\newblock \emph{ICLR}, 2016.

\bibitem[{Daniely} et~al.(2016){Daniely}, {Frostig}, and {Singer}]{daniely2016}
{Daniely}, A., {Frostig}, R., and {Singer}, Y.
\newblock {Toward Deeper Understanding of Neural Networks: The Power of
  Initialization and a Dual View on Expressivity}.
\newblock \emph{arXiv:1602.05897}, 2016.

\bibitem[Derrida \& Pomeau()Derrida and Pomeau]{derrida1986}
Derrida, B. and Pomeau, Y.
\newblock Random networks of automata: A simple annealed approximation.
\newblock \emph{EPL (Europhysics Letters)}, 1\penalty0 (2):\penalty0 45.

\bibitem[Elman(1990)]{elman1990finding}
Elman, Jeffrey~L.
\newblock Finding structure in time.
\newblock \emph{Cognitive science}, 14\penalty0 (2):\penalty0 179--211, 1990.

\bibitem[Glorot \& Bengio()Glorot and Bengio]{glorot2010understanding}
Glorot, Xavier and Bengio, Yoshua.
\newblock Understanding the difficulty of training deep feedforward neural
  networks.

\bibitem[Graves et~al.(2013)Graves, Mohamed, and Hinton]{graves2013speech}
Graves, Alex, Mohamed, Abdel-rahman, and Hinton, Geoffrey.
\newblock Speech recognition with deep recurrent neural networks.
\newblock In \emph{ICASSP}, pp.\  6645--6649. IEEE, 2013.

\bibitem[Greff et~al.(2017)Greff, Srivastava, Koutn{\'\i}k, Steunebrink, and
  Schmidhuber]{greff2017lstm}
Greff, Klaus, Srivastava, Rupesh~K, Koutn{\'\i}k, Jan, Steunebrink, Bas~R, and
  Schmidhuber, J{\"u}rgen.
\newblock Lstm: A search space odyssey.
\newblock \emph{IEEE transactions on neural networks and learning systems},
  28\penalty0 (10):\penalty0 2222--2232, 2017.

\bibitem[Haagerup \& Larsen(2000)Haagerup and Larsen]{haagerup2000brown}
Haagerup, Uffe and Larsen, Flemming.
\newblock Brown's spectral distribution measure for r-diagonal elements in
  finite von neumann algebras.
\newblock \emph{Journal of Functional Analysis}, 176\penalty0 (2):\penalty0
  331--367, 2000.

\bibitem[Hanin \& Rolnick(2018)Hanin and Rolnick]{hanin2018start}
Hanin, Boris and Rolnick, David.
\newblock How to start training: The effect of initialization and architecture.
\newblock \emph{arXiv preprint arXiv:1803.01719}, 2018.

\bibitem[Hayou et~al.(2018)Hayou, Doucet, and Rousseau]{hayou2018selection}
Hayou, Soufiane, Doucet, Arnaud, and Rousseau, Judith.
\newblock On the selection of initialization and activation function for deep
  neural networks.
\newblock \emph{arXiv preprint arXiv:1805.08266}, 2018.

\bibitem[{He} et~al.(2015){He}, {Zhang}, {Ren}, and {Sun}]{he2015}
{He}, K., {Zhang}, X., {Ren}, S., and {Sun}, J.
\newblock {Deep Residual Learning for Image Recognition}.
\newblock \emph{ArXiv e-prints}, December 2015.

\bibitem[Hidasi et~al.(2015)Hidasi, Karatzoglou, Baltrunas, and
  Tikk]{hidasi2015session}
Hidasi, Bal{\'a}zs, Karatzoglou, Alexandros, Baltrunas, Linas, and Tikk,
  Domonkos.
\newblock Session-based recommendations with recurrent neural networks.
\newblock \emph{arXiv preprint arXiv:1511.06939}, 2015.

\bibitem[Hochreiter \& Schmidhuber(1997)Hochreiter and
  Schmidhuber]{hochreiter1997long}
Hochreiter, Sepp and Schmidhuber, J{\"u}rgen.
\newblock Long short-term memory.
\newblock \emph{Neural computation}, 9\penalty0 (8):\penalty0 1735--1780, 1997.

\bibitem[Hyland \& R{\"a}tsch(2017)Hyland and R{\"a}tsch]{hyland2017learning}
Hyland, Stephanie~L and R{\"a}tsch, Gunnar.
\newblock Learning unitary operators with help from u (n).
\newblock In \emph{AAAI}, pp.\  2050--2058, 2017.

\bibitem[Ioffe \& Szegedy(2015{\natexlab{a}})Ioffe and Szegedy]{ioffe2015}
Ioffe, Sergey and Szegedy, Christian.
\newblock Batch normalization: Accelerating deep network training by reducing
  internal covariate shift.
\newblock In \emph{Proceedings of The 32nd International Conference on Machine
  Learning}, pp.\  448--456, 2015{\natexlab{a}}.

\bibitem[Ioffe \& Szegedy(2015{\natexlab{b}})Ioffe and Szegedy]{ioffe2015batch}
Ioffe, Sergey and Szegedy, Christian.
\newblock Batch normalization: Accelerating deep network training by reducing
  internal covariate shift.
\newblock In \emph{ICML}, pp.\  448--456, 2015{\natexlab{b}}.

\bibitem[Jozefowicz et~al.(2015)Jozefowicz, Zaremba, and
  Sutskever]{jozefowicz2015empirical}
Jozefowicz, Rafal, Zaremba, Wojciech, and Sutskever, Ilya.
\newblock An empirical exploration of recurrent network architectures.
\newblock In \emph{ICML}, pp.\  2342--2350, 2015.

\bibitem[Jozefowicz et~al.(2016)Jozefowicz, Vinyals, Schuster, Shazeer, and
  Wu]{jozefowicz2016exploring}
Jozefowicz, Rafal, Vinyals, Oriol, Schuster, Mike, Shazeer, Noam, and Wu,
  Yonghui.
\newblock Exploring the limits of language modeling.
\newblock \emph{arXiv preprint arXiv:1602.02410}, 2016.

\bibitem[{Karakida} et~al.(2018){Karakida}, {Akaho}, and {Amari}]{Karakida2018}
{Karakida}, R., {Akaho}, S., and {Amari}, S.-i.
\newblock {Universal Statistics of Fisher Information in Deep Neural Networks:
  Mean Field Approach}.
\newblock \emph{ArXiv e-prints}, June 2018.

\bibitem[Kiros et~al.(2015)Kiros, Zhu, Salakhutdinov, Zemel, Urtasun, Torralba,
  and Fidler]{kiros2015skip}
Kiros, Ryan, Zhu, Yukun, Salakhutdinov, Ruslan~R, Zemel, Richard, Urtasun,
  Raquel, Torralba, Antonio, and Fidler, Sanja.
\newblock Skip-thought vectors.
\newblock In \emph{Advances in neural information processing systems}, pp.\
  3294--3302, 2015.

\bibitem[Le et~al.(2015)Le, Jaitly, and Hinton]{le2015simple}
Le, Quoc~V, Jaitly, Navdeep, and Hinton, Geoffrey~E.
\newblock A simple way to initialize recurrent networks of rectified linear
  units.
\newblock \emph{arXiv preprint arXiv:1504.00941}, 2015.

\bibitem[LeCun et~al.(1998)LeCun, Bottou, Bengio, and
  Haffner]{lecun1998gradient}
LeCun, Yann, Bottou, L{\'e}on, Bengio, Yoshua, and Haffner, Patrick.
\newblock Gradient-based learning applied to document recognition.
\newblock \emph{Proceedings of the IEEE}, 86\penalty0 (11):\penalty0
  2278--2324, 1998.

\bibitem[Lee et~al.(2017)Lee, Bahri, Novak, Schoenholz, Pennington, and
  Sohl-Dickstein]{lee2017deep}
Lee, Jaehoon, Bahri, Yasaman, Novak, Roman, Schoenholz, Samuel~S, Pennington,
  Jeffrey, and Sohl-Dickstein, Jascha.
\newblock Deep neural networks as gaussian processes.
\newblock \emph{arXiv preprint arXiv:1711.00165}, 2017.

\bibitem[Marcus et~al.(1993)Marcus, Marcinkiewicz, and
  Santorini]{marcus1993building}
Marcus, Mitchell~P, Marcinkiewicz, Mary~Ann, and Santorini, Beatrice.
\newblock Building a large annotated corpus of english: The penn treebank.
\newblock \emph{Computational linguistics}, 19\penalty0 (2):\penalty0 313--330,
  1993.

\bibitem[Mikolov et~al.(2010)Mikolov, Karafi{\'a}t, Burget, Cernock{\`y}, and
  Khudanpur]{mikolov2010recurrent}
Mikolov, Tomas, Karafi{\'a}t, Martin, Burget, Lukas, Cernock{\`y}, Jan, and
  Khudanpur, Sanjeev.
\newblock Recurrent neural network based language model.
\newblock In \emph{Interspeech}, volume~2, pp.\ ~3, 2010.

\bibitem[Mishkin \& Matas(2015)Mishkin and Matas]{mishkin2015all}
Mishkin, Dmytro and Matas, Jiri.
\newblock All you need is a good init.
\newblock \emph{arXiv preprint arXiv:1511.06422}, 2015.

\bibitem[Neal(2012)]{neal2012}
Neal, Radford~M.
\newblock \emph{Bayesian learning for neural networks}, volume 118.
\newblock Springer Science \& Business Media, 2012.

\bibitem[Pascanu et~al.(2013)Pascanu, Mikolov, and
  Bengio]{pascanu2013difficulty}
Pascanu, Razvan, Mikolov, Tomas, and Bengio, Yoshua.
\newblock On the difficulty of training recurrent neural networks.
\newblock In \emph{International Conference on Machine Learning}, pp.\
  1310--1318, 2013.

\bibitem[Pennington et~al.(2017)Pennington, Schoenholz, and
  Ganguli]{pennington2017resurrecting}
Pennington, Jeffrey, Schoenholz, Sam, and Ganguli, Surya.
\newblock Resurrecting the sigmoid in deep learning through dynamical isometry:
  theory and practice.
\newblock \emph{NIPS}, 2017.

\bibitem[Pennington et~al.(2018)Pennington, Schoenholz, and
  Ganguli]{PenningtonSG18}
Pennington, Jeffrey, Schoenholz, Samuel~S., and Ganguli, Surya.
\newblock The emergence of spectral universality in deep networks.
\newblock In \emph{AISTATS}, pp.\  1924--1932, 2018.

\bibitem[{Poole} et~al.(2016){Poole}, {Lahiri}, {Raghu}, {Sohl-Dickstein}, and
  {Ganguli}]{poole2016}
{Poole}, B., {Lahiri}, S., {Raghu}, M., {Sohl-Dickstein}, J., and {Ganguli}, S.
\newblock {Exponential expressivity in deep neural networks through transient
  chaos}.
\newblock \emph{NIPS}, 2016.

\bibitem[Rumelhart et~al.(1986)Rumelhart, Hinton, and
  Williams]{rumelhart1986learning}
Rumelhart, David~E, Hinton, Geoffrey~E, and Williams, Ronald~J.
\newblock Learning representations by back-propagating errors.
\newblock \emph{nature}, 323\penalty0 (6088):\penalty0 533, 1986.

\bibitem[Saul et~al.(1996)Saul, Jaakkola, and Jordan]{saul1996}
Saul, Lawrence~K, Jaakkola, Tommi, and Jordan, Michael~I.
\newblock Mean field theory for sigmoid belief networks.
\newblock \emph{Journal of artificial intelligence research}, 4:\penalty0
  61--76, 1996.

\bibitem[Saxe et~al.(2013)Saxe, McClelland, and Ganguli]{saxe2013exact}
Saxe, Andrew~M, McClelland, James~L, and Ganguli, Surya.
\newblock Exact solutions to the nonlinear dynamics of learning in deep linear
  neural networks.
\newblock \emph{arXiv preprint arXiv:1312.6120}, 2013.

\bibitem[{Schoenholz} et~al.(2017){Schoenholz}, {Gilmer}, {Ganguli}, and
  {Sohl-Dickstein}]{schoenholz2016}
{Schoenholz}, S.~S., {Gilmer}, J., {Ganguli}, S., and {Sohl-Dickstein}, J.
\newblock {Deep Information Propagation}.
\newblock \emph{ICLR}, 2017.

\bibitem[Schoenholz et~al.(2017)Schoenholz, Pennington, and
  Sohl-Dickstein]{schoenholz2017correspondence}
Schoenholz, Samuel~S, Pennington, Jeffrey, and Sohl-Dickstein, Jascha.
\newblock A correspondence between random neural networks and statistical field
  theory.
\newblock \emph{arXiv preprint arXiv:1710.06570}, 2017.

\bibitem[Sompolinsky et~al.(1988)Sompolinsky, Crisanti, and
  Sommers]{sompolinsky1988}
Sompolinsky, H., Crisanti, A., and Sommers, H.~J.
\newblock Chaos in random neural networks.
\newblock \emph{Phys. Rev. Lett.}, 61:\penalty0 259--262, Jul 1988.
\newblock \doi{10.1103/PhysRevLett.61.259}.

\bibitem[Sussillo \& Abbott(2014)Sussillo and Abbott]{sussillo2014}
Sussillo, David and Abbott, LF.
\newblock Random walks: Training very deep nonlinear feed-forward networks with
  smart initialization.
\newblock \emph{CoRR, vol. abs/1412.6558}, 2014.

\bibitem[Tallec \& Ollivier(2018)Tallec and Ollivier]{tallec2018can}
Tallec, Corentin and Ollivier, Yann.
\newblock Can recurrent neural networks warp time?
\newblock \emph{arXiv preprint arXiv:1804.11188}, 2018.

\bibitem[Vorontsov et~al.(2017)Vorontsov, Trabelsi, Kadoury, and
  Pal]{vorontsov2017orthogonality}
Vorontsov, Eugene, Trabelsi, Chiheb, Kadoury, Samuel, and Pal, Chris.
\newblock On orthogonality and learning recurrent networks with long term
  dependencies.
\newblock \emph{arXiv preprint arXiv:1702.00071}, 2017.

\bibitem[Wisdom et~al.(2016)Wisdom, Powers, Hershey, Le~Roux, and
  Atlas]{wisdom2016full}
Wisdom, Scott, Powers, Thomas, Hershey, John, Le~Roux, Jonathan, and Atlas,
  Les.
\newblock Full-capacity unitary recurrent neural networks.
\newblock In \emph{Advances in Neural Information Processing Systems}, pp.\
  4880--4888, 2016.

\bibitem[Wu et~al.(2017)Wu, Ahmed, Beutel, Smola, and Jing]{wu2017recurrent}
Wu, Chao-Yuan, Ahmed, Amr, Beutel, Alex, Smola, Alexander~J, and Jing, How.
\newblock Recurrent recommender networks.
\newblock In \emph{Proceedings of the Tenth ACM International Conference on Web
  Search and Data Mining}, pp.\  495--503. ACM, 2017.

\bibitem[Xie et~al.(2017)Xie, Xiong, and Pu]{xie2017all}
Xie, Di, Xiong, Jiang, and Pu, Shiliang.
\newblock All you need is beyond a good init: Exploring better solution for
  training extremely deep convolutional neural networks with orthonormality and
  modulation.
\newblock \emph{arXiv preprint arXiv:1703.01827}, 2017.

\bibitem[Yang \& Schoenholz(2018)Yang and Schoenholz]{yang2018deep}
Yang, Greg and Schoenholz, Sam~S.
\newblock Deep mean field theory: Layerwise variance and width variation as
  methods to control gradient explosion, 2018.

\bibitem[Yang \& Schoenholz(2017)Yang and Schoenholz]{yang2017mean}
Yang, Greg and Schoenholz, Samuel~S.
\newblock Mean field residual networks: On the edge of chaos.
\newblock \emph{arXiv preprint arXiv:1712.08969}, 2017.

\bibitem[Zaremba et~al.(2014)Zaremba, Sutskever, and
  Vinyals]{zaremba2014recurrent}
Zaremba, Wojciech, Sutskever, Ilya, and Vinyals, Oriol.
\newblock Recurrent neural network regularization.
\newblock \emph{arXiv preprint arXiv:1409.2329}, 2014.

\bibitem[Zilly et~al.(2016)Zilly, Srivastava, Koutn{\'\i}k, and
  Schmidhuber]{zilly2016recurrent}
Zilly, Julian~Georg, Srivastava, Rupesh~Kumar, Koutn{\'\i}k, Jan, and
  Schmidhuber, J{\"u}rgen.
\newblock Recurrent highway networks.
\newblock \emph{arXiv preprint arXiv:1607.03474}, 2016.

\end{thebibliography}
\bibliographystyle{icml2018}

\onecolumn

\clearpage
\appendix

\normalsize
\part*{Supplemental material}

\setcounter{figure}{0} \renewcommand{\thefigure}{Supp.\arabic{figure}}
\setcounter{table}{0} \renewcommand{\thetable}{Supp.\arabic{table}}

\section{MinimalRNN Architecture}

\begin{figure}[!h]
\centering
    \includegraphics[width=0.4\textwidth]{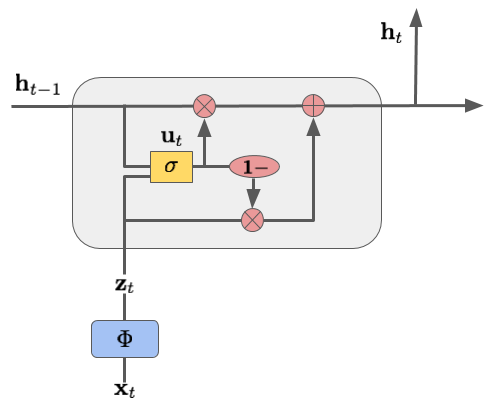}
\caption{Model architecture of minimalRNN.}
\label{fig:mrnn}
\end{figure}

\section{Diagonal Recurrence Relation}\label{supp:Q_recurrence}

Here we analyze the mean field dynamics of the minimalRNN. The minimalRNN features a hidden state $\bm h^t\in \mathbb R^N$ and inputs $\bm x^t$. The inputs are transformed via a fully-connected network $\bm z^t = \bm \Phi(\bm x^t)\in\mathbb R^M$ before being fed into the network. The RNN cell is then described by the equations,
\begin{align}
v^t_{i;a} &= \sum_j W_{ij} h^{t-1}_{j;a} + \sum_j V_{ij} z^t_{j;a} + b_i\\
u^t_{i;a} &= \sigma(v^t_{i;a})\\
h^t_{i;a} &= u^t_{i;a} h^{t-1}_{i;a} + (1-u^t_{i;a})z^t_{i;a}.
\end{align}
Here $i$ denotes the (pre)-activation and $a$ denotes an input to the network.Thus, $u^t_i$ acts as a gate on the $t$'th step. We take $W_{ij}\sim\mathcal N(0, \sigma_w^2/N)$, $V_{ij}\sim\mathcal N(0, \sigma_v^2/M)$ and $b_i\sim\mathcal N(\mu_b,\sigma_b^2)$.

By the CTL we can make a mean field assumption that $v^t_{i;a}\sim\mathcal N(\mu_b,q^t_{ab})$ where,
\begin{equation}
q^t_{ab} = \sigma_w^2\mathbb E[h^{t-1}_{i;a}h^{t-1}_{i;b}] + \sigma_v^2\mathbb E[z^t_{i;a}z^t_{i;b}] + \sigma_b^2 = \sigma_w^2Q_{ab}^{t-1} + \sigma_v^2 R_{ab}^t + \sigma_b^2
\end{equation}
where we have defined $Q_{ab}^t = \mathbb [h^t_{i;a}h^t_{i;b}]$ and $R_{ab}^t=\mathbb E[z^t_{i;a}z^t_{i;b}]$. We note that $R^t_{ab}$ is fixed by the input, but it remains for us to work out $Q^t_{ab}$. We find that,
\begin{align}
Q_{ab}^t = \mathbb E[h^t_{i;a}h^t_{i;b}] &= \mathbb E[u^t_{i;a}h^{t-1}_{i;a}u^t_{i;b}h^{t-1}_{i;b}] + \mathbb E[u^t_{i;a}h^{t-1}_{i;a}(1-u^t_{i;b})z^t_{i;b}] \\ &\hspace{2pc}+ \mathbb E[(1-u^t_{i;a})z^t_{i;a}u^t_{i;b}h^{t-1}_{i;b}] + \mathbb E[(1-u^t_{i;a})z^t_{i;a}(1-u^t_{i;b})z^t_{i;b}]\\
&\approx  \mathbb E[u_{i;a}^tu_{i;b}^t]\mathbb E[h^{t-1}_{i;a}h^{t-1}_{i;b}] + \mathbb E[(1-u_{i;a}^t)(1-u_{i;b}^t)]\mathbb E[z^t_{i;a}z^t_{i;b}]
\end{align}
where we have assumed that the expectation factorizes so that $h^{t-1}_{i;a}$ and $u^t_{i;a}$ are approximately independent. 

We choose to normalize the data so that $R_{aa}^t = R_{bb}^t = R$ independent of time. An immediate consequence of this normalization is that $Q_{aa}^t = Q_{bb}^t = Q^t$ and $q_{aa}^t = q_{bb}^t = q^t$. We then write $R_{ab}^t = R \Sigma^t$, $Q_{ab}^t = Q^tC^t$ and $q_{ab}^t = q^tc^t$ where $\Sigma^t$, $C^t$, and $c^t$ are cosine similarities between the inputs, the hidden states, and the $v^t_{a,b}$ respectively. With this normalization, we can work out the mean-field recurrence relation characterizing the covariance matrix for the minimalRNN.

We begin by considering the diagonal recurrence relations. We find that the dynamics are described by the equation, 
\begin{align}\label{eq:diagonal_mf_recurrence}
Q^t &= Q^{t-1}\int\mathcal Dz\sigma^2(\sqrt{q^t}z + \mu_b) + R\int\mathcal Dz\left[1-\sigma(\sqrt{q^t}z + \mu_b)\right]^2\\
q^t &=  \sigma_w^2Q^{t-1} + \sigma_v^2 R + \sigma_b^2
\end{align}
As expected, the first and second integrands determine how much of the update of the random network is controlled by the norm of the hidden state and how much is determined by the norm of the input. Since $\sigma(z) = 1-\sigma(-z)$ it follows that when $\mu_b = 0$ the first and second term will be equal and so,
\begin{equation}\label{eq:diagonal_mf_recurrence_zero_mean}
Q^t = (Q^{t-1} + R)\int\mathcal Dz\sigma^2(\sqrt{q^t}z).
\end{equation}
In general, $\mu_b$ will therefore control the degree to which the hidden state of the random minimalRNN is updated based on the previous hidden state or based on the inputs with $\mu_b = 0$ implying parity between the two. This is reflected in eq.~\eqref{eq:diagonal_mf_recurrence_zero_mean}.

\section{Existence of a $Q^*$ Fixed Point}
\label{supp:Q_fixed_point}

In the event that the norm of the inputs is time-independent, $R^t = R$ for all $t$, then the minimalRNN will have a fixed point provided there exists a $Q^*$ that satisfies a transcendental equation, namely that
\begin{align}\label{eq:diagonal_fixed_point}
\mathcal F(Q^*) \equiv \frac{\int\mathcal D_{q^*}z\left[1-\sigma\left(z\right)\right]^2}{\int\mathcal D_{q^*}z\left[1 - \sigma^2\left(z\right)\right]} - \frac{Q^*}{R} = 0\,.
\end{align}
It is easy to see that such a solution always exists. When $Q^*\to\infty$ the first term of $\mathcal F(Q^*)$ approaches $1$ while the magnitude of the second increases without bound and so $\mathcal F(Q^*) < 0$. Conversely, when $Q^*\to0$ the first term is positive while $Q^*/R\to 0$ and so $\mathcal F(Q^*) > 0$. The existence of a $Q^*$ satisfying the transcendental equation then follows directly from the intermediate value theorem. 

\section{$Q^*$ Dynamics}\label{supp:Q_dynamics}

We can now investigate the dynamics of the norm of the hidden state in the vicinity of $Q^*$. To do this suppose that $Q^t = Q^* + \epsilon^t$ with $\epsilon\ll 1$. Our goal is then to expand eq.\eqref{eq:diagonal_mf_recurrence} about $Q^*$. First, we note that,
\begin{align}
\sigma(\sqrt{q^t}z + \mu_b) &= \sigma(\sqrt{q^* + \sigma_w^2\epsilon^t}z + \mu_b)\\
&\approx \sigma\left(\sqrt{q^*}z + \mu_b +  \frac1{2\sqrt{q^*}}\sigma_w^2\epsilon^tz\right)\\
&\approx \sigma\left(\sqrt{q^*}z + \mu_b\right) +  \frac1{2\sqrt{q^*}}\sigma_w^2\epsilon^tz\sigma'\left(\sqrt{q^*}z + \mu_b\right) + \mathcal O((\epsilon^t)^2).
\end{align}
Letting $\zeta(z) = \sqrt{q^*}z + \mu_b$ this implies that,
\begin{align}
Q^t &= Q^{t-1}\int\mathcal Dz\sigma^2(\sqrt{q^* + \sigma_w^2\epsilon^{t-1}}z + \mu_b) + R\int\mathcal Dz\left[1-\sigma(\sqrt{q^* + \sigma_w^2\epsilon^{t-1}}z + \mu_b)\right]^2\\
Q^* + \epsilon^t &=Q^*\int\mathcal Dz\sigma^2\left(\zeta(z)\right) + R\int\mathcal Dz\left[1-\sigma(\zeta(z))\right]^2 + \epsilon^{t-1}\bigg[\int\mathcal Dz\sigma^2\left(\zeta(z)\right) \\&\hspace{3pc}+ \frac{Q^*}{\sqrt{q^*}}\sigma_w^2\int\mathcal Dzz\sigma(\zeta(z))\sigma'(\zeta(z)) 
- R\sqrt{q^*}\sigma_w^2\int\mathcal Dzz(1-\sigma(\zeta(z)))\sigma'(\zeta(z))\bigg]\\
\epsilon^t &=\epsilon^{t-1}\bigg[\int\mathcal Dz\sigma^2\left(\zeta(z)\right) + \frac{Q^*}{\sqrt{q^*}}\sigma_w^2\int\mathcal Dzz\sigma(\zeta(z))\sigma'(\zeta(z)) 
- R\sqrt{q^*}\sigma_w^2\int\mathcal Dzz(1-\sigma(\zeta(z)))\sigma'(\zeta(z))\bigg]\\
&=\epsilon^{t-1}\int\mathcal Dz\left[\sigma^2(\zeta(z)) + \frac{\sigma_w^2}{\sqrt{q^*}}\left\{(Q^*+R)\sigma(\zeta(z)) - R\right\}z\sigma'(\zeta(z))\right]\\
&=\epsilon^{t-1}\int\mathcal Dz\left[\sigma^2(\zeta(z)) + \frac{\sigma_w^2}{\sqrt{q^*}}\left\{(Q^*+R)\sigma(\zeta(z)) - R\right\}z\sigma(\zeta(z))(1-\sigma(\zeta(z)))\right]\\
&=\epsilon^{t-1}\int\mathcal Dz\left[\sigma^2(\zeta(z)) + \frac{\sigma_w^2}{\sqrt{q^*}}\left\{(Q^*+R)\sigma(\zeta(z)) - R\right\}z\sigma(\zeta(z))(1-\sigma(\zeta(z)))\right]
\end{align}
It follows that $q^t\to q^*$ as,
\begin{equation}
|q^t - q^*| \sim e^{-t/\xi_Q}
\end{equation}
with 
\begin{equation}
\xi^{-1}_Q = -\log\left(\int\mathcal Dz\left[\sigma^2(\zeta(z)) + \frac{\sigma_w^2}{\sqrt{q^*}}\left\{(Q^*+R)\sigma(\zeta(z)) - R\right\}\sigma'(\zeta(z))\right]\right)
\end{equation}
as expected. 

\section{Off-Diagonal Recurrence Relation}\label{supp:C_recurrence}

We now turn our attention to the off-diagonal term. From eq.~\eqref{eq:minimal_rnn_expectation} it follows that,
\begin{align}\label{eq:supp_minimal_rnn_off_diagonal_raw}
&Q^tC^t = Q^{t-1}C^{t-1}\int\mathcal Dz_1\mathcal Dz_2\sigma(u_1^t)\sigma(u_2^t) + R\Sigma^t \int\mathcal Dz_1\mathcal Dz_2 (1-\sigma(u_1^t))(1-\sigma(u_2^t))\\
&q^tc^t = \sigma_w^2 Q^{t-1}C^{t-1} +  \sigma_v^2R\Sigma^t + \sigma_b^2
\end{align}
where 
\begin{equation}
u_1^t = \sqrt{q^t}z_1 + \mu_b \hspace{1pc}\text{and}\hspace{1pc} u_2^t = \sqrt{q^t}\left(c^t z_1 + \sqrt{1-(c^t)^2}z_2\right) + \mu_b.
\end{equation}
By expanding eq~\eqref{eq:supp_minimal_rnn_off_diagonal_raw} as $c^t = c^* + \epsilon^t$ we find $\epsilon^{t+1} = \chi_{c^*}\epsilon^t$ where,
\begin{equation}
\chi_{c^*} = \int\mathcal Dz_1\mathcal D z_2\sigma(u_1)\sigma(u_2) + q^*(c^* + J_-)\int\mathcal Dz_1\mathcal Dz_2\sigma'(u_1)\sigma'(u_2).
\end{equation}
We note that when $c^* = 1$ it follows that $\chi_{c^*} = \chi_1$.

\section{Additional Hyperparameter Ranges}
\label{supp:ptb}
We tune the learning hyper-parameters in the following ranges for all the models:
\begin{itemize}[noitemsep]
\item learning rate: \{0.1, 0.2, 0.3, 0.5, 1, 2\}
\item max-epoch: \{4, 7, 11\}
\item decay: \{0.5, 0.65, 0.8\}
\item dropout: \{0.0, 0.2, 0.3, 0.5\}
\end{itemize}

\end{document}